# NFTGAN: Non-Fungible Token Art Generation Using Generative Adversarial Networks

SAKIB SHAHRIAR

College of Technological Innovation, Zayed University, United Arab Emirates

KADHIM HAYAWI[*]

College of Technological Innovation, Zayed University, United Arab Emirates

Digital arts have gained an unprecedented level of popularity with the emergence of non-fungible tokens (NFTs). NFTs are cryptographic assets that are stored on blockchain networks and represent a digital certificate of ownership that cannot be forged. NFTs can be incorporated into a smart contract which allows the owner to benefit from a future sale percentage. While digital art producers can benefit immensely with NFTs, their production is time consuming. Therefore, this paper explores the possibility of using generative adversarial networks (GANs) for automatic generation of digital arts. GANs are deep learning architectures that are widely and effectively used for synthesis of audio, images, and video contents. However, their application to NFT arts have been limited. In this paper, a GAN-based architecture is implemented and evaluated for novel NFT-style digital arts generation. Results from the qualitative case study indicate that the generated artworks are comparable to the real samples in terms of being interesting and inspiring and they were judged to be more innovative than real samples.

**CCS CONCEPTS** • Computing methodologies • Machine learning • Machine learning approaches

**Additional Keywords and Phrases:** NFT, blockchain, deep learning, adversarial networks, digital art generation

## 1 INTRODUCTION

Arts have always played a fundamental role in human lives across cultures and civilizations. It allows humans to express their thoughts, emotions, and memories. As the digital revolution continued with the introduction of smart phones and increase in internet accessibility, a demand for expressing arts in the digital world and social media was created. Although research in computer generated arts have been ongoing since the mid-twentieth century [1], it is after the digital revolution and social media adoption that their demand from a consumer perspective is at the highest. However, a major challenge for digital arts was the issue of ownership and copyright. Unlike physical arts, digital arts can be easily copied and distributed causing their value to significantly degrade. This obstacle is overcome with the introduction of non-fungible tokens (NFTs). NFTs are cryptographic tokens that were initially built on the Ethereum blockchain but more recently adapted by other blockchains including Solana [2]. NFTs are different from fungible tokens like Bitcoins, where each token is similar in value to the other ones and can easily be swapped [3]. The non-fungible aspect of NFTs ensures that each token is unique and consequently provides the holder with ownership rights over a digital asset.

[*] Corresponding Author email: Abdul.Hayawi@zu.ac.ae
Address: College of Technological Innovation, Zayed University, Abu Dhabi Campus, MF2-0-018
Contact Number: +9712 599 3545

Production and sales of digital arts skyrocketed with the introduction of NFTs. 'Everydays – The First 5000 Days', an NFT art created by the artist Beeple sold for a record-breaking 69.3 million United State dollars during March of 2021 [4]. NFTs provide several major benefits to artists including the fact that artists no longer have to transfer ownership and content to agents and that artists can benefit by programming NFTs to receive a predetermined loyalty fee for any future transactions of their artworks [5]. Moreover with the emergence of metaverse, a virtual environment blending physical and digital realities [6], NFT-based digital art is set to become a core component in providing many metaverse experiences [7]. This has caught the attention of artists around the globe in producing digital arts. Younger artists are particularly interested and engaged in "crypto art" [8]. However, digital art production is a time consuming process and requires extensive knowledge of computer software and programming. Therefore, automatic digital art creation is highly desirable and one technology enabling automatic art generation is generative adversarial networks (GANs).

GANs are deep learning architectures that are typically made up of two neural network-based models known as generators and discriminators. The generator network is trained to synthesize new data points from random noise and the discriminator is trained to recognize between the original and generated data points. The two models are trained together in a zero-sum game and ideally the discriminator will be fooled by the generator about half of the time, at which point the real and generated examples are virtually indistinguishable. GANs have produced state-of-the-art performance in various applications including text generation [9] and text to image generation [10]. Other related applications include the use of GANs in handwriting generation with calligraphic styles [11] and photorealistic images of video game generation [12]. Although software and computing approaches were traditionally used for digital art generation, the ability of GANs to generate arts automatically without human supervision makes them ideal for digital art generation [13]. This is revolutionary because once the network has been trained, it can synthesize a large quantity of very diverse set of artworks. It is then possible for the artist to become the "judge" and hand-pick the most inspiring artwork and perhaps enhance it further. For a comprehensive review of GAN-based art generation including visual arts, audio arts, and literary texts, the readers are encouraged to refer to the following survey paper [13]. Although widely and effectively used for various art-related applications, GANs are yet to be experimented with the task of NFT art generation. Therefore, this paper presents a novel NFT art generation application using GANs. Following are the main contributions of this paper:

- It presents a novel NFT art generation application trained using GANs.
- It provides a quantitative evaluation and a qualitative case study of NFT arts generated using GANs.
- It discusses the impacts, limitations, and future research work in GAN-based NFT art generation.

The rest of the paper is organized as follows. Section two outlines the methodology used for NFT art generation and the evaluation approaches. The results are presented and discussed in Section three. Section four discusses the implications, limitations, and future work. Section five concludes this paper.

## 2  METHODS

In this section, the overall methodology including the dataset used and preprocessing steps, the network architecture, and evaluation metrics are discussed.

### 2.1  Dataset and Preprocessing

A publicly available dataset on Kaggle (https://www.kaggle.com/vepnar/nft-art-dataset) was utilized for the proposed application. The dataset contains NFT arts in gif, image, and video file formats. For the purpose of this work, only the image files were utilized. The image dataset contains 3021 NFT-style art images. The art images were not labeled to a specific category, and they represent a very diverse selection of NFT arts.

All the images were resized to 512 by 512 size which is compatible with the GAN architecture. Although this led to the GAN training time being much longer, the expected art quality increased significantly. Moreover, any images that were non-RGB color format were removed. After this step, a total of 2283 images were left in the final dataset to be used for training.

### 2.2  GAN Architecture and Experimental Setup

There are various GAN architectures available for image generation including deep convolutional GANs, style transfer models, and conditional GANs [14]. Conditional GANs are suitable for generating data points based on a label or condition. Given that the dataset is not labeled, the objective in this work is to generate the arts unconditionally. As such, the GAN architecture would be trained to learn the overall NFT art style from the dataset and consequently generate novel arts based on the existing style. A popular style transfer-based model known as StyleGAN [15] was employed for the proposed NFT arts generation application. In comparison to other models, StyleGAN is capable of generating high definition face images using two mechanisms, i.e., generating high-resolution images using progressive growing [16] and using adaptive instance normalization [17] to infuse image styles in each layer. The main difference between a traditional generator and the one utilized in StyleGAN is the addition of a network that enables mapping from the latent space to a high dimensional space before adding fully connected layers to another network that synthesizes content with input noise. StyleGAN 2 [18] was introduced as an improvement to the original architecture, which often produced unnecessary artifacts or effects in generated images. The removal of the unwanted effects was obtained by restructuring the adaptive instance normalization mechanism to a weight demodulation network, which scales the convolutional weights to perform suitable style transfer.

The StyleGAN2 [19] version, which is suitable for smaller datasets, was utilized for this work. The *PyTorch* [20] library implementation for StyleGAN2 was used. The model training was performed on Google Colab pro using the NVIDIA Tesla P100 GPU with 16 GB memory. The model was trained for about 59 hours with checkpoints after every five iterations. The model pickle files were saved, and training was resumed after about 10 hours due to time limit restrictions on the platform.

## 2.3 Evaluation

After the training phase, the generated NFT artworks were evaluated on quantitative and qualitative metrics. For the quantitative evaluation, a popular metric known as Fréchet Inception Distance (FID) [21] was utilized, which is defined in Equation 1.

$$FID = \|\mu_r - \mu_g\|^2 + T_r\left(\Sigma_r + \Sigma_g - 2(\Sigma_r\Sigma_g)^{1/2}\right) \quad (1)$$

where $\mu_r$ and $\mu_g$ represents the feature-wise mean of the real and the generated samples respectively, $T_r$ is the trace operation, and $\Sigma_r$ and $\Sigma_g$ represents the covariance matrix for the real and generated feature vector respectively.

In addition to the FID score, kernel inception distance (KID) was also used for evaluation. KID was proposed by [22] as an extension to FID score using a polynomial kernel that eliminates the potential bias of FID scores in certain situations. KID is computed using the squared maximum mean discrepancy and KID scores are asymptotically normal. For both FID and KID metrics, a lower score indicates better performance.

Furthermore, to assess the generated NFT arts quality, a qualitative case study was conducted. A total of 26 university students and graduates were invited to evaluate 6 of the generated samples and 6 of the samples from the dataset. Following the survey, the results were aggregated and analyzed. For each of the 12 images, the volunteers were asked to answer the following questions:

- Do you find the artwork interesting? (1 very uninteresting, 5 very interesting)
- Do you find the artwork inspiring? (1 very uninspiring, 5 very inspiring)
- Do you think the artwork is innovative? (1 extremely unoriginal, 5 very innovative)
- Rate the art quality overall. (1 poor, 5 excellent)
- Do you think the art work is created by an artist or a computer?

## 3 RESULTS

Training progress was monitored using the saved model after every 5 iterations. Moreover, the FID score was also examined to assess the training progress. Figure 1 depicts the training progress starting from the initialization step until the final iteration. The model starts off by producing random coloured noises, then moves on to creating consistent shapes, then progressively produces more diverse images, and then finally generates the desired artworks.

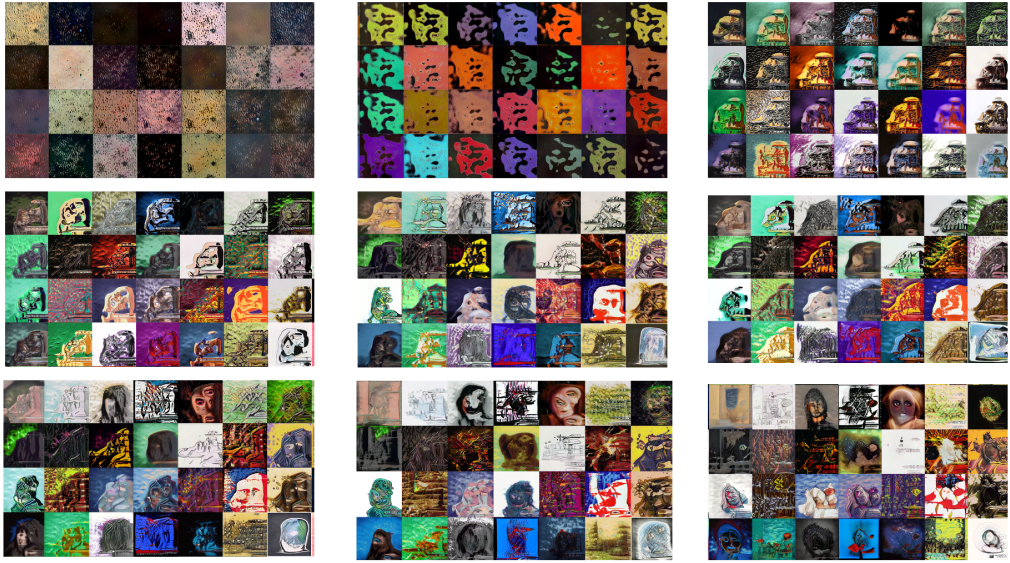

Figure 1 Training progress in generating artworks

After approximately 59 hours of training, the improvement in FID score was negligible and consequently the training was stopped, and the best model was saved. Figure 2 illustrates some of the generated samples. The generated arts were diverse in terms of their style and color. A compilation of the generated arts in video format is also available online (https://youtu.be/THcYhDk9f-Y). Next, the evaluation results are presented.

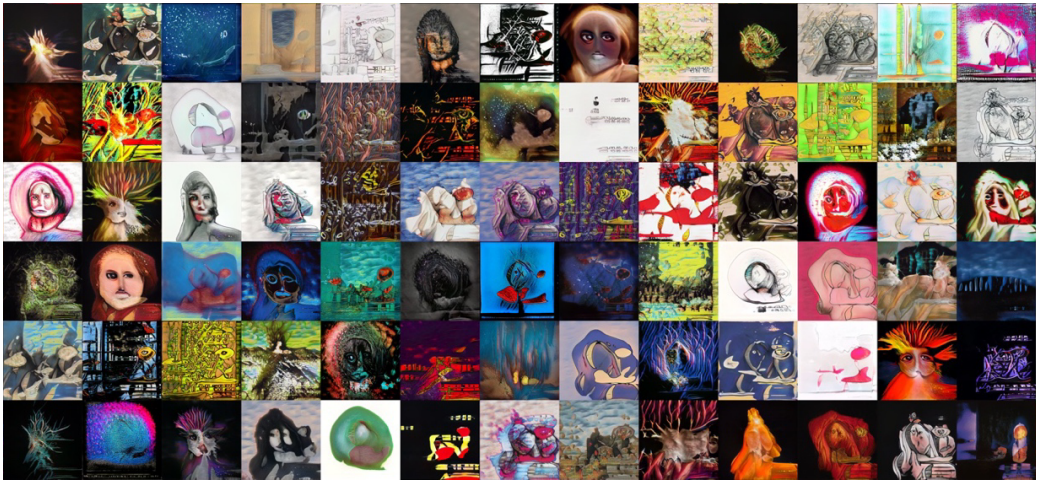

Figure 2 GAN-generated NFT arts

### 3.1 Quantitative Results

In this section, the results using the aforementioned FID and KID metrics are presented. Although the FID score was monitored during the training phase, the last model was used for final evaluation. Table 1 summarizes the scores from the best model.

Table 1: Results using evaluation metrics

| Metric | FID | KID |
|---|---|---|
| Results | 43.64 | 0.012 |

The low scores for both FID and KID indicate that the model was able to produce high quality generations. Overall, the results obtained using StyleGAN in this work outperformed some of the related applications in the literature that reported using the same metrics [11], [12]. However, it must be noted that these metrics are generally used to compare performance across various GAN architectures solving a related generation task. In such cases, these metrics can help in deciding the best performing model.

### 3.2 Quantitative Results

The results of the qualitative case study from the 26 volunteers for the four different criteria are summarized in Table 2. The table indicates the results for both the real artworks from the dataset and the GAN generated artworks. Each evaluation criteria is scored between 1 to 5 (1 indicating poor quality and 5 indicating excellent quality). The average scores along with their standard deviation is presented in Table 2.

Table 2: Results using qualitative case study

| Evaluation Criteria | Real Samples | Generated Samples |
|---|---|---|
| Interesting | 3.68 ± 0.32 | 3.46 ± 0.06 |
| Inspiring | 3.35 ± 0.31 | 3.10 ± 0.11 |
| Innovative | 3.35 ± 0.30 | 3.47 ± 0.09 |
| Overall | 3.35 ± 0.26 | 3.24 ± 0.16 |

The results of the case study indicate that the generated samples were close to the real samples for the four evaluation characteristics. The real samples were superior in terms of being more interesting, inspiring and, their overall quality. Interestingly, the volunteers deemed the GAN generated artworks to be more innovative. Moreover, the volunteers were also asked if they thought the artworks were produced by artists or were computer generated. For the GAN generated samples, the volunteers believed the generated samples to be created by real artists 40.6% of the time. This indicates that many of the generated samples were of very high quality which confused the evaluators from determining their origin. Furthermore, the standard deviations for the generated samples are low indicating that most of the evaluators scored similarly and there appears to be a consensus. Despite the limited samples size, the results from the case study indicate that GAN-based NFT artwork generation is promising. The results from the case study are displayed in Figure 3.

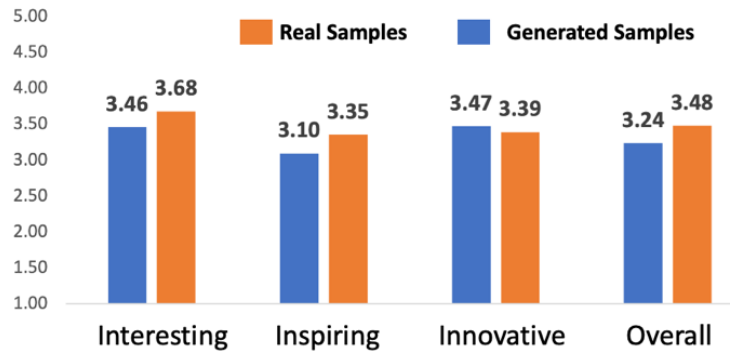

Figure 3: Case study results comparison between real and generated artworks

## 4 DISCUSSION

The results in the previous section demonstrate the immense potential of GAN-based solutions for NFT arts. Based on the results, further discussions are presented next.

### 4.1 Implications

The proposed application has several key implications. Firstly, with the growing interest in blockchain and NFT technologies, this work will facilitate further research interest in GAN-based solutions in NFT related applications. This will lead to experimentation with different forms of digital arts generation and facilitate the introduction of newer datasets and algorithms. Moreover, the encouraging results obtained in this work highlight the outstanding potential of GANs in NFT art generation. This would consequently attract frequent collaborations between artists and machine learning engineers in producing high quality digital arts. Furthermore, the proposed work will significantly reduce the time and resources needed by professional artists to create digital arts. The artists can simply use the GAN generated arts that match their style and enhance them based on their requirements instead of creating digital arts from scratch. NFTs are also indicated to have diversification benefits during unsettled times including the ongoing COVID-19 pandemic [23]. This is significant for artists as they do not lose substantial value of their works even during unstable market conditions.

### 4.2 Limitations

The proposed work has several limitations. Firstly, all the arts were generated using a single GAN architecture, which is a style transfer-based model. Although the results were encouraging, the full potential of GANs in NFT art generation is not realized with a single model. It is possible that other models may outperform the StyleGAN on this application. Moreover, the qualitative case study was conducted on a smaller scale with a limited sample size consisting of only 26 volunteers. Therefore, the results obtained cannot be used to draw any firm conclusions. Finally, specialists and expert artists were not invited for evaluating the generated arts. Therefore, the reliability and effectiveness of the case study is limited.

### 4.3 Future Work

In this work, the NFT art generation was unconditional because the artworks were not labeled based on a category. To improve on the generation quality, it would be reasonable to experiment with conditional art generation focusing on a specific art content or style. For instance, a dataset of only animal arts could be used to train a GAN to generate arts of a specific animal such as cats or dogs conditionally. To realize this kind of application, there is a need for quality data collection which would initially require some collaboration with artists and professionals. Moreover, unconditional art generation would also compel experimentation with different GAN architectures including conditional GANs and cycle GANs as opposed to the sole architecture used in the proposed work. Besides artworks in image format, NFT arts also include artworks in GIFs and video file formats. As such, researchers are encouraged to take on the challenge of generating arts in other file formats including GIFs and short videos. Finally, a comprehensive case study with a bigger sample size including professional artists should be carried out to evaluate the potential of GANs in generating NFT arts.

## 5 CONCLUSION

In this work, a novel NFT art generation application using GANs was presented. A style transfer-based model, StyleGAN2, was trained on a publicly available collection of NFT arts dataset. The generated arts were of high visual quality and scored well on two metrics, FID and KID. Moreover, volunteers were invited to assess the quality of the arts and they found the arts to be comparable to the real samples from the dataset. The generated artworks scored 3.24 overall out of 5 in comparison to the slightly higher overall score of 3.35 out of 5 for the real samples. The proposed work has the potential to attract further research on GAN related NFT applications. Moreover, future research directions were highlighted including conditional generation of NFT arts as well as NFT art generation in other file formats such as GIFs and videos.

### ACKNOWLEDGMENTS

This work was supported by Zayed University RIF grant R20132.